\documentclass[10pt,twocolumn,letterpaper]{article}

\usepackage{wacv}
\usepackage{times}
\usepackage{epsfig}
\usepackage{graphicx}
\usepackage{amsmath}
\usepackage{amssymb}

\newcommand{\comment}[1]{}

\makeatletter
\@namedef{ver@everyshi.sty}{}
\makeatother
\usepackage{todonotes}

\usepackage{url}
\usepackage{float}
\usepackage{xspace}

\usepackage[newitem,newenum]{paralist}

\wacvfinalcopy
\def\VEC#1{{\boldsymbol{#1}}}

\makeatletter
\DeclareRobustCommand\onedot{\futurelet\@let@token\@onedot}
\def\@onedot{\ifx\@let@token.\else.\null\fi\xspace}
\def\eg{\emph{e.g}\onedot} 
\def\ie{\emph{i.e}\onedot}

\def\wrt{w.r.t\onedot} 
\def\etal{\emph{et al}\onedot}
\makeatother

\renewcommand{\eqref}[1]{Eq\onedot\ref{#1}}
\newcommand{\eqsref}[1]{Eqs.\,\ref{#1}}
\newcommand{\figref}[1]{Fig.\,\ref{#1}}

\newcommand{\secref}[1]{Sec.\,\ref{#1}}

\usepackage[pagebackref=true,breaklinks=true,letterpaper=true,colorlinks,bookmarks=false]{hyperref}

\def\wacvPaperID{567} 

\ifwacvfinal\fi
\begin{document}

\title{Appearance and Shape from Water Reflection}

\author{Ryo Kawahara \and Meng-Yu Jennifer Kuo \and Shohei Nobuhara \and Ko Nishino 
\and 
Kyoto University, Kyoto, Japan \\
{\tt\small http://vision.ist.i.kyoto-u.ac.jp/}
}

\maketitle
\ifwacvfinal\thispagestyle{empty}\fi

\begin{abstract}
This paper introduces single-image geometric and appearance reconstruction from water reflection photography, \ie, images capturing direct and water-reflected real-world scenes. Water reflection offers an additional viewpoint to the direct sight, collectively forming a stereo pair. The water-reflected scene, however, includes internally scattered and reflected environmental illumination in addition to the scene radiance, which precludes direct stereo matching. We derive a principled iterative method that disentangles this scene radiometry and geometry for reconstructing 3D scene structure as well as its high-dynamic range appearance. In the presence of waves, we simultaneously recover the wave geometry as surface normal perturbations of the water surface. Most important, we show that the water reflection enables calibration of the camera. In other words, for the first time, we show that capturing a direct and water-reflected scene in a single exposure forms a self-calibrating HDR catadioptric stereo camera. We demonstrate our method on a number of images taken in the wild. The results demonstrate a new means for leveraging this accidental catadioptric camera.
\end{abstract}


\section{Introduction}

Water reflection has long been a source of artistic inspiration. Various paintings come to mind that compose reflection by a water surface together with direct sight of a scene, such as Claude Monet's Autumn in Argenteuil. Water reflection has also been an integral part of architectural design as seen in Taj Mahal and Matsumoto Castle to name a few. Water reflection has also been used as an artistic expression in modern photography, for instance, by capturing a cityscape reflected in a puddle. 

It is perhaps much less understood that water reflection carries visual cues for scene structure recovery. A computer vision researcher, however, would likely notice that water reflection would give a different vantage point of the scene from the camera viewpoint when captured in a single image, collectively forming a (flipped) stereo pair. This suggests an opportunity for single-image scene geometry recovery. In fact, Yang et al. \cite{YangTIP_15} applied standard stereo reconstruction to estimate scene depth from a single water reflection image after adapting reflected scene appearance to construct a cost volume robust to their radiometric distortions. \comment{It, however, turns out that this 3D reconstruction is non-trivial. The views are not only geometrically different, but they are also radiometrically distorted.}

\begin{figure}[t]
 \centering
 \includegraphics[width=\linewidth]{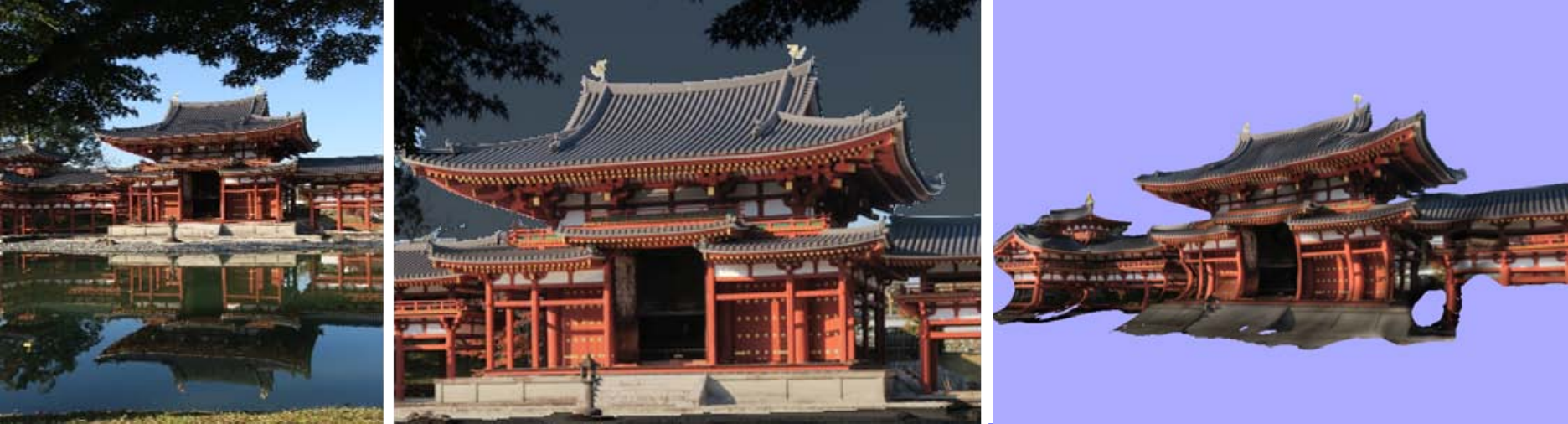}
    \caption{We show that we can recover the 3D geometry (right) and high-dynamic range appearance (middle) of a scene from a single image (left) capturing it both directly and through reflection by a water surface.}
 \label{fig:opening}
\end{figure}

In this paper, for the first time to our knowledge, we show that a single image capturing both the direct and reflected observation through water reflection of a scene results from a self-calibrating high-dynamic range catadioptric imaging system. That is, in sharp contrast to merely leveraging the geometric configuration of water reflection, we show that a high-dynamic range appearance and 3D shape of the scene can be recovered without any knowledge about the image formation a priori.\comment{formulate and solve \textit{shape from water reflection}, \ie, 3D reconstruction of a scene from a single image capturing it in both direct sight and reflection by a water surface.} We first consider the case where the water surface is calm and can be modeled as a planar mirror. As shown in \figref{fig:opening}, we derive a method that recovers high-dynamic range appearance and 3D structure of the scene. The main challenge lies in the fact that water-reflected scene radiance is compounded with environmental light scattered in the water medium and also reflected by the bottom surface. Scene radiance must be sifted out from this superposition in order to match against the direct observation for triangulation. In other words, radiometry and geometry recovery are inherently intertwined. 

We derive a canonical iterative method to recover scene geometry from the direct and water-reflected observations. We also show that high-dynamic range scene radiance can be estimated in the process and water reflection even enables calibration of the camera. That is, we do not need to know anything about the camera; its intrinsic parameters can be recovered by seeking agreements in angular-dependent Fresnel effects in the reflected observation, and its extrinsic parameters can be estimated by identifying the water surface. In other words, we show that capturing direct and water-reflected scenes in a single exposure forms a self-calibrating HDR catadioptric stereo camera.

%
The water surface in the image is not always calm and can have waves that lead to noticeable displacements in the reflected observation. We show that this can be modeled as surface normal variations of the water surface and derive an iterative approach to simultaneously recover the shape of both the scene structure and the water surface. For this, we introduce a principled method for incorporating realistic prior knowledge such as piecewise planarity for scene geometry and a Fourier-domain wave representation.

We experimentally validate our method both quantitatively and qualitatively by reconstructing scene structure and radiance from a wide range of real images. The results agree with the newly derived theory and demonstrate the effectiveness of its application to arbitrary images taken in the wild. The proposed method provides a new means for visually appreciating our 3D world, and enables a new form of 2D-3D visual media of \textit{water reflection photography}.

\section{Related works}

To our knowledge, the work by Yang et al. \cite{YangTIP_15} is the only other work that recovers 3D from an image of water reflection. This work modifies a standard stereo algorithm to compute two cost volumes, one for direct and another for reflected views, whose filtered disparities are later fused to produce a depth map. The method uses automatically established keypoint pairs between the direct and reflected views to adapt the appearance and also limit the range of disparities. This inevitably necessitates the automatically detected keypoints to uniformly span both the spatial and depth variations of the scene, which hinders the method's applicability--their results are all on well-textured natural scenes. The radiometric distortions in the reflected view are assumed to be rectifiable with simple linear adaptation both locally and globally. This assumption is simply incorrect due to the compound angular-dependent mixture of light as we later show and model. In contrast, instead of treating them as nuisance that needs to be corrected, we exploit the unique radiometric properties of water reflection as a rich source for true scene appearance recovery and show that it also enables self-calibration of the camera. That is, we show how to recover not just the geometry but also the radiometry of a scene, which are inherently intertwined, from a single water reflection image.

Appearance and shape from water reflection can be interpreted as accidental catadioptric stereo imaging in which the water surface serves as the catoptric view that forms a stereo pair with the direct dioptric view. Examples of accidental imaging include the use of occluders as pinspeck (anti-pinhole) cameras that capture surroundings as shadows\,\cite{torralba2012accidental}, which can be used to estimate the scene behind the occluders\,\cite{bouman2017turning,baradad2018inferring}. Accidental micro motions due to, for example, heart beating, can provide scene depth cues that can be used for image refocusing and synthetic parallax generation\,\cite{yu14three,im2015depth,ha2016high}. 

Catadioptric imaging\,\cite{baker1999theory,Sturm2008} in computer vision has been applied to a variety of tasks including omnidirectional imaging\,\cite{scaramuzza2006flexible}, reflectance acquisition\,\cite{mukaigawa2011hemispherical}, shape-from-silhouette\,\cite{forbes06shape}, structured light\,\cite{lanman09surround,tahara15interference}, kaleidoscopic imaging\,\cite{reshetouski11three,takahashi2017linear}, and stereo reconstruction\,\cite{adelson1992single,nane98stereo,somanath2010single}. Gluckman and Nayer\,\cite{gluckman2001catadioptric}, in particular, proposed a catadioptric stereo system with two planar mirrors. As the first example of an accidental catadioptric system, Nishino and Nayar\,\cite{KonCVPR04,nishino2006corneal} showed that capturing eyes form a catadioptric imaging system in which the cornea serves as the reflector.

Stereo matching with translucency\,\cite{tsin2003stereo,xiong2009fractional} or image-based reflection separation\,\cite{levin2004user,sarel2004separating,kong2014physically,SHWARTZ20082164} explicitly model transmission through semi-translucent surfaces. They utilize either 3D recovery or models in the Fourier domain for blind separation of reflected and transmitted images. They cannot, however, be applied to non-planar surfaces such as wavy water surfaces.


\section{Assumptions}

Let us first clarify the assumptions we make. As we consider an image capturing both a direct view of a scene and its reflection by a water surface, we can safely make the following assumptions without loss of generality.
\begin{itemize}
 \item The water medium (\eg, pond or puddle) is homogeneous and has a known index of refraction.
    \item The reflected scene as well as the bottom of the water medium consist of Lambertian surfaces. 
    \item The reflection is purely specular at the water surface. 
    \item The sun is not captured in the reflection.
    \item We can either manually or automatically isolate the image region capturing water reflection.
\end{itemize}

We do not require knowledge of the camera parameters neither intrinsic nor extrinsic. If EXIF information is available, we use it to initialize the intrinsic camera parameters. We show that these camera parameters can be estimated from the image. 

When the water surface has waves, we assume that they satisfy the following properties.
\begin{itemize}
 \item The wave amplitude (\ie, height) is comparatively smaller than the camera height from the water surface.
 \item Any interreflection on the water surface is negligible in intensity and one water surface point reflects one (but not necessarily unique) object surface point.
\end{itemize}

\section{Planar Water Reflection}

\begin{figure}[t]
 \centering
 \includegraphics[width=\linewidth]{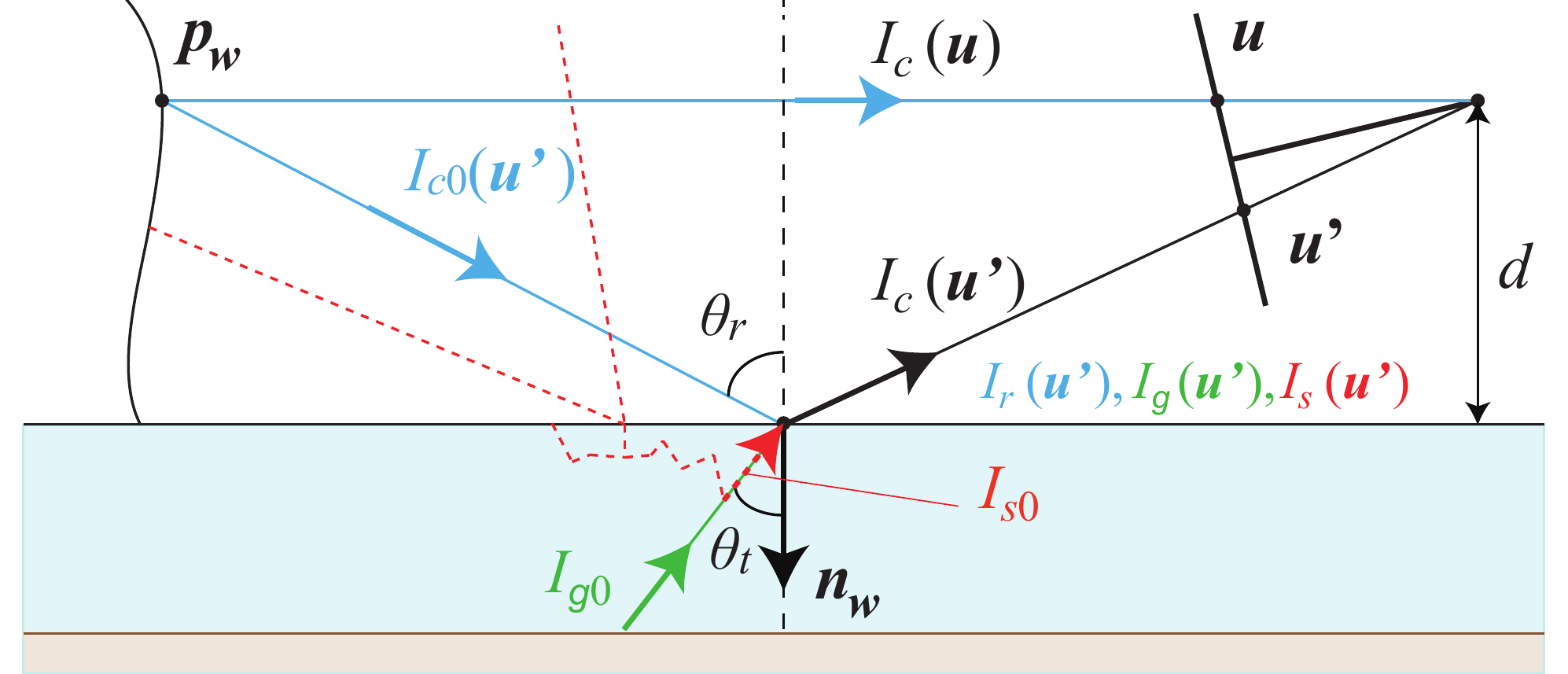}
 \caption{A scene point $\VEC{p}_w$ is observed directly ($\VEC{u}$) and also through reflection $\VEC{u}'$ by the water surface. Unlike the direct observation $I_c(\VEC{u})$ the reflected radiance $I_c(\VEC{u}')$ contains not just the direct radiance from the scene point $I_{c0}(\VEC{u}')$ which gets specularly reflected by the water surface, but also the radiance of light that has scattering through the water medium $I_{s0}$ and that reflected off the bottom surface of it $I_{g0}$ that is refracted into the camera.}
 \label{fig:model_intensity}
\end{figure}

We begin by deriving the key steps for recovering appearance and shape from planar water reflection which are also shared when dealing with wavy water reflection. As depicted in \figref{fig:model_intensity}, a 3D scene point $\VEC{p}_w$ is observed twice in the image: the direct observation $I_c(\VEC{u})$ and the reflected observation $I_c(\VEC{u}')$. Note that both \textit{observations} of scene points are captured in a single image and represented by their positions on the image plane, $\VEC{u}=(u_x,U_y,1)^\top$ and $\VEC{u}'=({u'}_x,{U'}_y,1)^\top$, respectively. 




\subsection{Planar Water Surface Reconstruction}\label{sec:planarwater}

For a calm water surface, or as an initial estimate of the global surface normal of a wavy water surface, we estimate the surface normal of the water surface $\VEC{n}_w$ from a small set of corresponding pairs of direct and reflected observations $(\VEC{u}, \VEC{u}')$ satisfying
\begin{equation}
	\begin{split}
    \VEC{u}'^\top A^{-\top} 
	[\VEC{n}_{w}]_\times A^{-1} \VEC{u}' &= 0\,,
	\label{eq:get_n}
	\end{split}
\end{equation}
where $A$ is the intrinsic parameter matrix of the camera and $[\VEC{n}_{w}]_\times$ is the skew-symmetric matrix of $\VEC{n}_w$. This is a linear constraint on the normal $\VEC{n}_w$, and we can estimate $\VEC{n}_w$ from two or more corresponding pairs. Intrinsic camera parameters $A$ can be initialized with EXIF information, when available, or with reasonable values common in outdoor photography, which are then refined by the radiometric recovery process as we detail in \secref{sec:intrinsics}.

We obtain these correspondences semi-automatically. We first segment the image into direct and reflected observation regions, which can often be done by just specifying the line where the direct and reflected observations meet in the image. We then run generic feature detection and matching methods in these two regions. For a calm water surface, we found this process to be sufficient for all cases. For a moderately wavy water surface, we conduct this automatic feature matching on a downsampled and blurred image to obtain the global surface normal of the water surface. Note that we only need a few correspondences as we are only recovering the water surface normal.

\subsection{Direct--Reflected Stereo Reconstruction}\label{sec:stereorecon}

For a calm water surface that can be modeled as a planar reflector, once we estimate its normal, the direct and reflected observations in the image form a stereo image pair (albeit folded). For a wavy water surface, the stereo correspondence pairs are locally perturbed by the varying surface normal at each water surface point. In either case, if the displacements due to waves are undone and correspondences are established as we later show, we may recover the 3D coordinates of scene points via regular stereo reconstruction (\ie, triangulation) from the direct--reflected observation point pairs:
\begin{align}
    \begin{cases}
	\VEC{u}  &= \lambda_c A \VEC{p}_w\,,  \\
	\VEC{u}' &= \lambda'_c A ( H_w \VEC{p}_w + \VEC{t}_w)\
    \end{cases}
\Leftrightarrow
	\begin{pmatrix}
	u_x M_3-M_1 \\ u_y M_3-M_2 \\
	{u'}_x M'_3-M'_1 \\ 
	{u'}_y M'_3-M'_2 
	\end{pmatrix} \VEC{p}_w = 0,
	\label{eq:projection}
\end{align}
where $H_w = (I - 2\VEC{n}_w\VEC{n}_w^\top)$ is a householder matrix, and $\VEC{t}_w = 2 d \VEC{n}_w$. $M=(A \ \VEC{0})$, $M'=A(H_w \  \VEC{t}_w)$ are projection matrices for each viewpoint.

This stereo reconstruction requires the intrinsic and extrinsic parameters of the camera. The extrinsic camera parameters can be described by the mirrored camera pose $H$ and translation $\VEC{t}_w$ \wrt the original viewpoint defined by the water surface normal $\VEC{n}_w$ and distance $d$ of the camera from the water surface as depicted in \figref{fig:model_intensity}. Since the global scale is not known, we assume $d=1$ and the recovered scene geometry is scaled accordingly. 


\section{Appearance from Water Reflection}\label{sec:appearance}

As we show in \figref{fig:model_intensity}, the reflected observation of a scene point $I_c(\VEC{u}')$ is a superposition of specular reflection by the water surface $I_r(\VEC{u}')$ of the Lambertian scene radiance $I_{c0}(\VEC{u}')$, environmental light scattered through the water medium $I_s(\VEC{u}')$, and also bouncing off the bottom surface $I_g(\VEC{u}')$
\begin{equation}
	\begin{split}
		I_c(\VEC{u}) &= I_{c0}(\VEC{u}')\,, \\
		I_c(\VEC{u}') &= I_r(\VEC{u}') + I_g(\VEC{u}') + I_s(\VEC{u}')\,.
	\end{split}
	\label{eq:intensity_camera}
\end{equation}
We need to separate these components and recover the scene radiance $I_{c0}(\VEC{u}')$, so that correspondences can be established between the reflected and direct observations. 

\subsection{Reflected Scene Radiance}
The reflected scene radiance by the water surface $I_r$ can be described by the Fresnel power reflectance $F$ 
\begin{equation}
		I_r(\VEC{u}') = F(\VEC{u}') I_{c0}(\VEC{u}')\,.
		\label{eq:intensity_reflect}
\end{equation}
The power reflectance $F$ under natural light is
\begin{equation}
		F(\VEC{u}') = \frac{1}{2} (R_s+R_p)\,, 
		\label{eq:fresnel}
\end{equation}
where $R_s$ and $R_p$ are the power reflection coefficients for s-polarized and p-polarized light given respectively by
\begin{equation}
		R_s = \left( \frac{ \sin(\theta_r - \theta_t) }{ \sin(\theta_r + \theta_t) } \right)^2\,,\quad 
		R_p = \left( \frac{ \tan(\theta_r - \theta_t) }{ \tan(\theta_r + \theta_t) } \right)^2\,.
		\label{eq:fresnel_polarized}
\end{equation}

The angle of incidence $\theta_r$, which is also the angle of specular reflection, is 
\begin{equation}
    \theta_r = \cos^{-1}\left( (A^{-1} \VEC{u}')^{\top} \VEC{n}_w \right)\,,
    \label{eq:theta_r}
\end{equation}
and the angle of refraction $\theta_t$ becomes
\begin{equation}
        \theta_t = \sin^{-1} \left( \frac{\mu_a}{\mu_w} \sin(\theta_r) \right)\,,
        \label{eq:theta_t}
\end{equation}
from Snell's law. $\mu_a$, $\mu_w$ are the refraction indices of the air and water respectively.

\subsection{Water-Scattered Environmental Illumination}

We assume that the environmental illumination is uniform across the water surface, \ie, the incident environmental illumination to the water surface points in the reflected-observation region does not vary. This is a reasonable assumption as long as the sun is not directly imaged via water reflection. Part of this environmental light is transmitted into water, scattered in the medium, and then transmitted back into the viewpoint.  

The scattered environmental illumination is the sum of scattered light from all points along the transmitted path observed at $\VEC{u}'$. On the other hand, the bottom of the water medium at which the light path intersects will have a comparatively weaker radiance, which suggests that we can limit the contribution of water-scattered environmental illumination to that from near the water surface. We model this near-surface water-scattered environmental illumination based on the dipole method\,\cite{Jensen2001dipole}, which describes the process of environmental illumination 
$I(\theta_i)$ from $\theta_i$ transmitted into water with $T(\theta_i) = 1 - F(\theta_i)$, attenuated with $R_d(\tau)$, then transmitted again into the camera $T(\theta_r(\VEC{u}'))$ for all angle $\theta_i$ and distance $\tau$:
\begin{equation}
	\begin{split}
		I_s(\VEC{u}') & = \int_{\tau} \int_{\theta_i}  T(\theta_i ) R_d(\tau) T(\theta_r(\VEC{u}')) I(\theta_i)d\tau d{\theta_i}\,, \\
		& \simeq T( \theta_r(\VEC{u}') ) I_{s0}\,, 
	\end{split}
	\label{eq:intensity_scattered}
\end{equation}
where $I_{s0}$ is the homogeneous scattered radiance. Note that, for water, the BSSRDF is simply $T(\theta_i ) R_d(\tau) T(\theta_r(\VEC{u}'))$, since the scattering attenuation only depends on the distance $\tau$ between the incident and exitant surface points.

Note that, when the medium is shallow, the bottom surface reflection will be comparatively strong and instead be transmitted into the viewing direction almost without any scattering. This effect will be accounted for by the bottom surface reflection model described in \secref{sec:bottom}. 

\subsection{Reflection From The Bottom Surface}\label{sec:bottom}

The reflected observation also includes environmental illumination reflected by the bottom surface of the water medium. Since the incident environmental illumination would be scattered to and from the water bottom, and as we assume a Lambertian bottom surface, this light can be denoted as 
\begin{equation}
		I_g(\VEC{u}') = T(\theta_r(\VEC{u}')) I_{g0}\,,
		\label{eq:intensity_ground}
\end{equation}
where $I_{g0}$ is the uniform Lambertian reflection at the bottom, and $T(\theta_r(\VEC{u}'))$, or abbreviated as $T(\VEC{u}')$, is the transmittance into the camera.

It is important to note that this component is negligible when considering reflected observations by ponds and lakes that have sufficient depth. It becomes dominant only for shallow water media such as puddles.

\subsection{Scene Radiance and HDR Recovery}\label{sec:hdr}

\begin{figure}[t]
 \centering
 
  \includegraphics[width=0.5\linewidth]{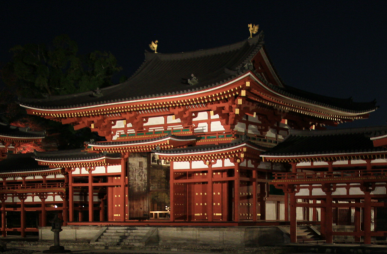}
  \includegraphics[width=0.49\linewidth]{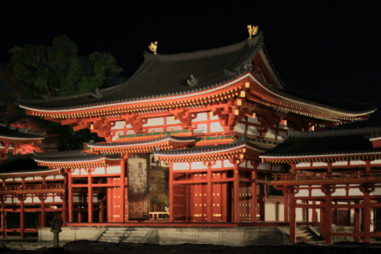}
 
 \caption{The Fresnel reflection on the water surface, in effect, provides scene radiance captured with spatially varying exposures that are different from the camera. By combining these two radiance exposures, we reconstruct high-dynamic range appearance of the scene (right), which reveals scene details in saturated regions in the original direct view (left).}
 \label{fig:hdr}
\end{figure}

To recover the scene radiance from the reflected observation, we first estimate the sum of diffuse bottom surface reflectance $I_{g0}$ and subsurface scattering $I_{s0}$ from each of $N$ pairs of direct and reflected image coordinates $(\VEC{u}_i, \VEC{u}'_i)$ and their observations $( I_c(\VEC{u}_i), I_c(\VEC{u}'_i) )\;(i=1,\dots,N)$ and average them using \eqsref{eq:intensity_reflect}, \ref{eq:intensity_scattered}, \ref{eq:intensity_ground} 
\begin{equation}
	\begin{split}
		I_c(\VEC{u}'_i) &= F(\VEC{u}'_i) I_c(\VEC{u}_i) + T(\VEC{u}'_i) (I_{g0} + I_{s0})\,, \\
		I_{g0} + I_{s0} &= \frac{1}{N} \sum_{i=1}^{N} \left\{ T(\VEC{u}'_i)^{-1} \left(I_c(\VEC{u}'_i) - F(\VEC{u}'_i) I_c(\VEC{u}_i) \right) \right\} \,.
	\end{split}
	\label{eq:bgcolor_estimation}
\end{equation}

We can then sift out the scene radiance from the reflected observation of each $\VEC{u}'$ by subtracting these additive components and by undoing Fresnel reflection
\begin{equation}
		I_{c0}(\VEC{u}') = F(\VEC{u}')^{-1} \left( I_c(\VEC{u}') - T(\VEC{u}') (I_{g0} + I_{s0}) \right) \,.
	\label{eq:Ic0_estimation}
\end{equation}
The recovered scene radiance in the reflected observation is then used for stereo matching with its direct observation.

The reflected observation consists of a darkened scene radiance due to Fresnel reflection combined with scattered and bottom-reflected environmental illumination. The latter components are usually less dominant compared to the Fresnel-reflected scene radiance component. Once we eliminate those components, we are basically left with the scene radiance modulated by the angular-varying Fresnel effect, which reduces the scene radiance more as we look at closer water surface in a general image capture setting. In other words, we are left with the scene radiance captured with a varying neutral density filter, which suggests that we can combine the scene radiance values captured in the direct and reflected observations to estimate high-dynamic range appearance of the scene. Although this is HDR recovery from only two different exposures, as shown in \figref{fig:hdr}, it lets us recover scene appearance details particularly in saturated regions of the direct view or underexposed regions of the reflected view. In \secref{sec:exps}, we demonstrate this HDR scene recovery and show tone-mapped results\,\cite{Banterle17hdr}.


\subsection{Intrinsic Camera Parameter Estimation}\label{sec:intrinsics}



In regular stereo reconstruction, changes in intrinsic camera parameters do not alter the fundamental matrix, which in turns means that the camera intrinsics cannot be recovered. In contrast, in shape from water reflection, the reflected scene radiance is altered by the Fresnel water surface reflection whose magnitude depends on the viewing angle $\theta_r(\VEC{u}')$ (equivalently the incident angle). This suggests that we may estimate intrinsic camera parameters, most notably the focal length $f_c$, when recovering the reflected scene radiance from the reflected observation.

\eqref{eq:theta_r} indicates that the viewing angle $\theta_r$ is also a function of the intrinsic camera matrix $A$. The focal length $f_c$ together with the environmental illumination components $I_{g0}+I_{s0}$ can be estimated by minimizing errors of \eqsref{eq:bgcolor_estimation} and \ref{eq:Ic0_estimation} 
\begin{align}
  &\mathop{\rm arg\,min}\limits_{ f_c,I_{g0}+I_{s0} } \sum_{ \langle\VEC{u}, \VEC{u}' \rangle } \left( 
  E_{c0} + \lambda E_{gt} \right) \,, \\
  &E_{c0} = \| I_{c0}(f_c, \VEC{u}', I_{g0}+I_{s0}) - I_c(\VEC{u}) \| \,, \\ 
  &E_{gt} = \| I_{g0}+I_{s0} -  T(\VEC{u}')^{-1} \left(I_c(\VEC{u}') - F(\VEC{u}') I_c(\VEC{u}) \right) \| \,,
\end{align}
where $\lambda$ is a weighting parameter. The second term $E_{gt}$ evaluates the uniformity of the estimated environmental illumination.



\section{Wavy Water Reflection}
\begin{figure}[t]
 \centering
 \includegraphics[width=\linewidth]{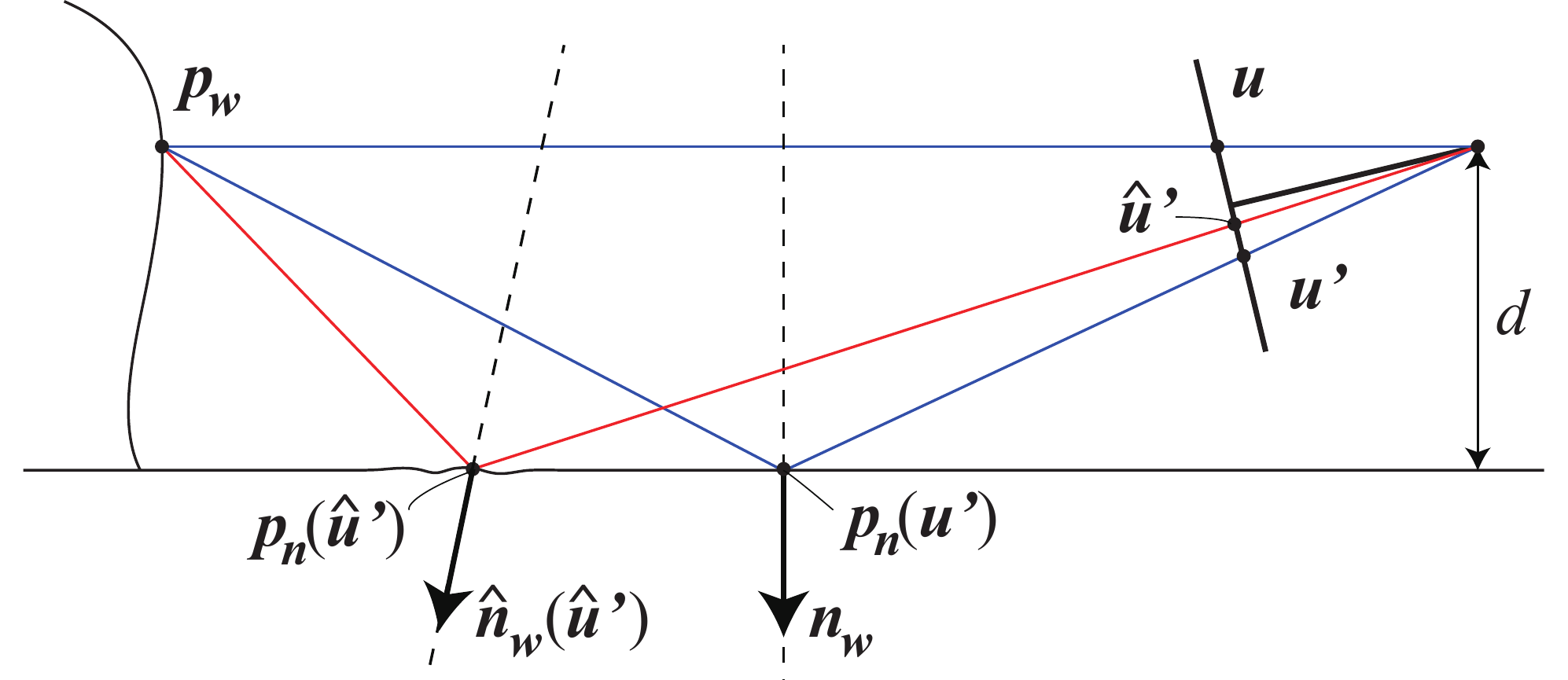}
 \caption{By estimating waves as surface normal variations of the water surface, we can remove the effects of waves and recover the reflection point and image coordinates of the reflected observation for a planar water surface.}
 \label{fig:model_wave_normal}
\end{figure}
When the water surface has noticeable waves, we simultaneously estimate the geometry of the waves and the scene.

\subsection{Wavy Water Surface}
As we can safely assume that the wave amplitude is sufficiently small compared to the height of the camera, we can model them as surface normal perturbations to the otherwise flat water surface. This surface normal variation causes changes in projected image coordinates and their radiance (\ie, reflected observations). Using notations depicted in \figref{fig:model_wave_normal}, we model this by expressing the reflected observation for a corresponding pair of direct and reflected observations $(\VEC{u}, \VEC{\hat{u}'})$ using the local normal $\VEC{\hat{n}}_w(\VEC{\hat{u}'})$ of the water surface point from where that reflected observation comes
\begin{equation}
	\begin{split}
		I_c(\VEC{\hat{u}'}) = & F(\VEC{\hat{u}'}, \VEC{\hat{n}}_w(\VEC{\hat{u}'})) I_c(\VEC{u}) \\
		& + T(\VEC{\hat{u}'}, \VEC{\hat{n}}_w(\VEC{\hat{u}'})) (I_{g0} + I_{s0})\,.
	\end{split}
	\label{eq:intensity_camera_wave}
\end{equation}



We model the reflected observation as that taken by a collection of reflected viewpoints, \ie, pixel-wise mirrored cameras with mirrored poses $H_w(\VEC{\hat{u}'})$ and translations $\VEC{t}_w(\VEC{\hat{u}'})$ 
\begin{equation}
	\VEC{\hat{u}'} = \lambda'_c A \left( 
	            H_w(\VEC{\hat{u}'}) \VEC{p}_w + \VEC{t}_w(\VEC{\hat{u}'})
	            \right)\,,
	\label{eq:projection_wave}
\end{equation}
which makes explicit the relationship of the 3D image coordinates $\VEC{u}$ and $\VEC{\hat{u}'}$ in terms of a sum of deformation and disparity on the 2D image plane.

\begin{figure*}[t]
 \centering
 \includegraphics[width=\linewidth]{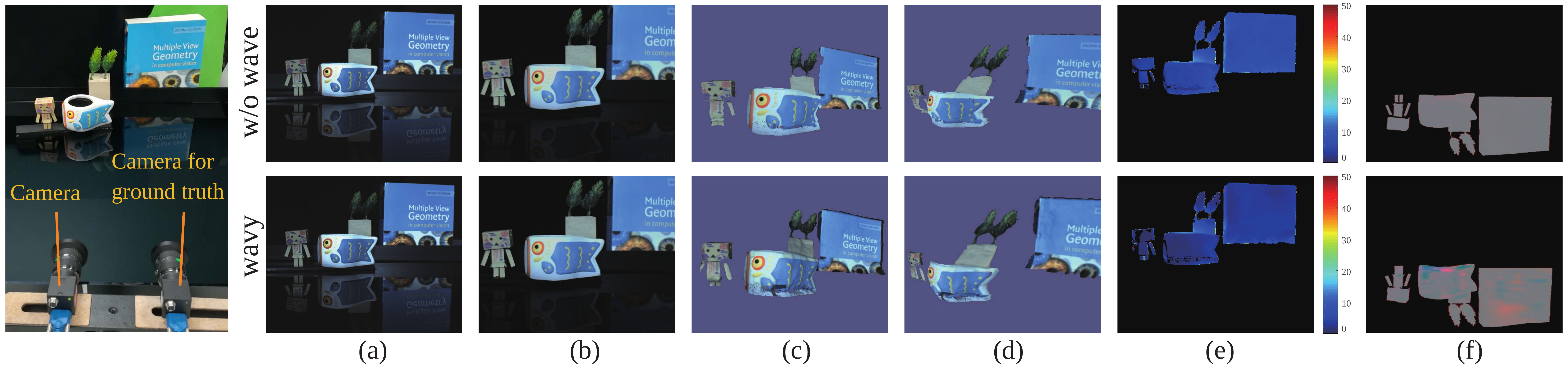}
 \caption{Quantitative evaluation using structured-light stereo to acquire ground truth depth. From (a) a single image with bottom surface reflection and waves, we recover (b) HDR appearance and (c,d) dense 3D geometry (shown in two views) which agree well with ground truth as (e) the percentile error map \wrt scene depth range shows. Our method also recovers the wave structure as surface normal variations as shown with (f) ten-times amplified normal maps. }
 \label{fig:quant_eval}
\end{figure*}

\subsection{Waves as 2D Deformations}\label{sec:2dreg}

Our goal is to remove the effects of water surface deformation and associated radiometric change in reflected observation (\ie, image intensity change) by comparing direct and reflected observations independent of the disparity. The main challenge for this is that the angular dependency of the reflected observation only results in a subtle change in image intensity which, in general, is not large enough to robustly separate the water surface normal variation and disparity. To simultaneously estimate the wavy water surface and the scene geometry, we formulate it as a 2D image alignment and employ prior knowledge about the scene structure and the wavy water surface.

We first project the direct and reflected observations to a common image plane on which we can achieve 2D image alignment between the observations. While we may use arbitrary virtual planes, for simplicity, we use the water surface seen fronto-parallel as the common image plane. Let us denote the common image plane with $\{\VEC{u}'\}$. Note that the 3D image coordinates $\{\VEC{u}'\}$ are unknown because we only know $I_c(\VEC{u})$ and $I_c(\VEC{\hat{u}'})$. We denote the direct observations in the actual input image with $I_D(\VEC{u}')$ and those projected onto the common image plane with $I_c(\VEC{u})$. Similarly, we denote the wave-removed reflected observation (\ie, the estimated reflected scene appearance for a planar water surface) with $I_M(\VEC{u}')$ and its recovered radiometry (\ie, estimated scene radiance) with $I_c(\VEC{\hat{u}'})$.

Our objective is to estimate the waves as local surface normals of the water surface such that 
\begin{equation}
	  I_D(\VEC{u}') = I_M(\VEC{u}')\,. 
\end{equation}
The geometric projection of 3D image coordinates of the direct observation $\VEC{u}$ to that on the common image plane $\VEC{u}'$ in $I_D$ is 
\begin{equation}
	 \VEC{u}' = H_{disp}(\VEC{p}_w) \VEC{u}\,,  
	 \label{eq:u_to_u'}
\end{equation}
where $H_{disp}(\VEC{p}_w)$ is a homography matrix determined by the 3D scene point $\VEC{p}_w$ and the local water surface normal $\VEC{n}_w$ for each pixel $\VEC{u}$.

On the other hand, the geometric transformation 3D image coordinates of the reflected observation $\VEC{\hat{u}'}$ to that on the common image plane $\VEC{u}'$ in $I_M$ is 
\begin{equation}
	  \VEC{u}' = g( \VEC{\hat{u}'} )\,.
	  \label{eq:hatu'_to_u'}
\end{equation}
We estimate this displacement field $g$ with a generic 2D non-rigid image registration method\,\cite{Vercauteren09demon}.

\subsection{Wavy Water Surface Reconstruction}\label{sec:waverecon}
The estimated displacement field on the common image plane provides correspondences between direct and reflected observations $(\VEC{u}, \VEC{u}')$ and the water surface normals $\VEC{n}_w$ and scene geometry $\VEC{p}_w$ can be recovered from \eqref{eq:u_to_u'}. 
As depicted in \figref{fig:model_wave_normal}, the displacement estimate tells us that the reflected observation $\VEC{\hat{u}'}$ is moved to $\VEC{u}'$ on the input image for the same scene point $\VEC{p}_w$. The corresponding reflection point on the water surface is moved to $\VEC{\VEC{p}_n(\VEC{u}')}$ after removing the wave 
\begin{equation}
	 \VEC{p}_n(\VEC{u}') = \frac{d}{\cos(\theta_r(\VEC{u}'))} \VEC{v}_c(\VEC{u}')\,.
	 \label{eq:p_n}
\end{equation}

By using the estimated 3D scene coordinates $\VEC{p}_w$, we obtain the surface normal at each reflecting point of the wavy water 
\begin{equation}
    \VEC{\hat{n}}_w(\VEC{\hat{u}'}) = \frac{1}{2} \left(
        -\frac{\VEC{p}_n(\VEC{\hat{u}'})}{|\VEC{p}_n(\VEC{\hat{u}'})|} + \frac{\VEC{p}_w - \VEC{p}_n(\VEC{\hat{u}'})}{|\VEC{p}_w-\VEC{p}_n(\VEC{\hat{u}'})|}\,.
	\right)
	\label{eq:hatn_w}
\end{equation}

When the waves are erroneously recovered, the recovered scene geometry will subsume the errors in water surface geometry resulting in a wavy 3D scene structure. Since real-world surface geometry is, in general, not wave-like, we can impose a geometric prior on the scene. We employ a piecewise planar geometry prior, that encourages the recovered scene geometry to consist of locally planar surfaces. In particular, we segment the direct observation of the scene into superpixels and impose this piecewise planarity on each superpixel (\ie, a locally connected set of $\{\VEC{u}\}$). We impose this prior in a coarse-to-fine fashion, in which the superpixel size is iteratively refined. This, in effect, allows smoothly curved scene geometry while nudging the wave pattern to be explained by the water surface instead of the scene structure.

Since 2D non-rigid image registration on the common image plane can also erroneously absorb disparity errors as surface normal variations, we also impose a prior on the wave geometry. Inspired by simple computer graphics models for water waves\,\cite{Darles11wave, Li18wave}, we employ a Fourier domain prior model. The height map $h(\VEC{p}_n, t_0 )$ for a 3D wavy water surface point $\VEC{p}_n$ at a single time instance is given as
\begin{equation}
	 h(\VEC{p}_n ) = \VEC{p}_0 + \sum_{i} a_i e^{i\VEC{k} \VEC{p}_n -\phi_i}\,,
\end{equation}
where $a_i$ is the amplitude of a Fourier component, and $\VEC{k}$ is a wave vector. The surface normals of the wavy water surface is computed by differentiation of the wave heights. We use a pre-determined number of Fourier components and apply inverse Fourier transform to recover the wave.

%

\section{Experimental Results}\label{sec:exps}
\begin{figure}[t]
 \centering
 \includegraphics[trim={0 1cm 0 0.5cm},clip ,width=\linewidth]{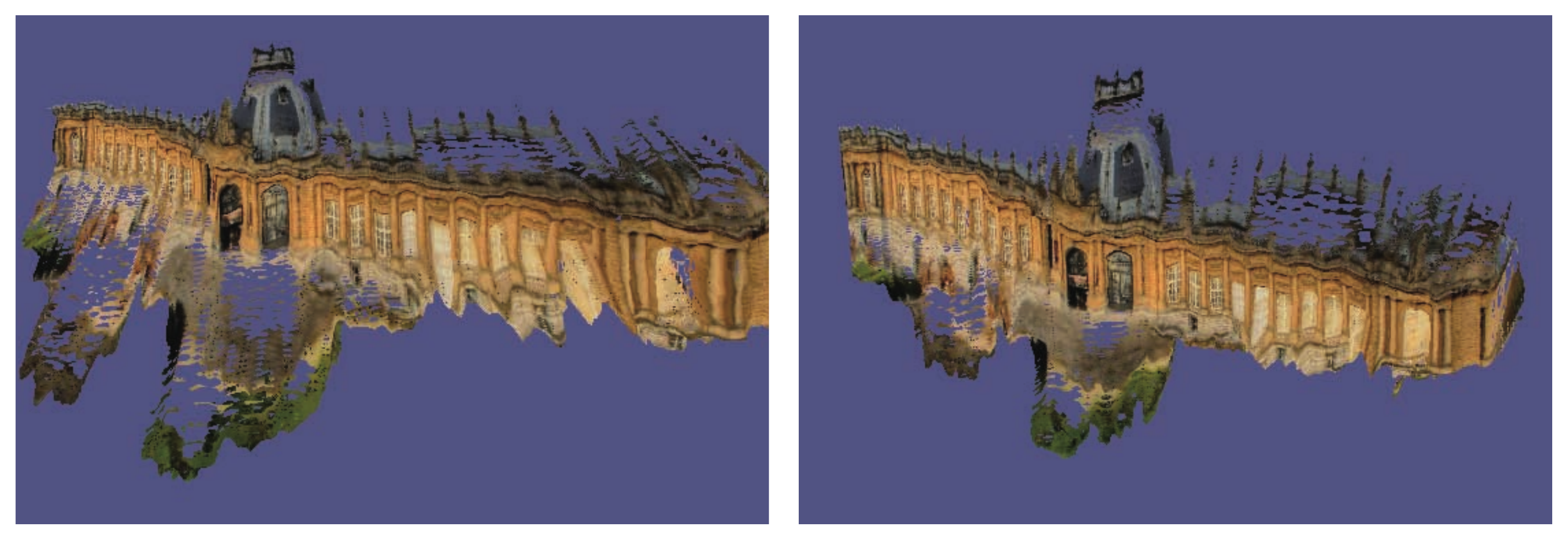}
 \caption{The effect of self-calibration: reconstructed 3D model without (left) and with (right) focal length estimation. The simultaneous focal length estimation (\ie, self-calibration) clearly results in more accurate geometry.}
 \label{fig:estimate_intrinsics}
\end{figure}

We thoroughly evaluate and demonstrate the effectiveness of our method with controlled experiments and with arbitrary images taken in the wild.

To quantitatively evaluate the accuracy of the recovered 3D scene geometry, as shown in the left most of \figref{fig:quant_eval}, we setup a dual-camera imaging system in the lab. Although only one camera is used for capturing the input image for our method, we use the other camera and a projector to reconstruct ground truth depth with structured light stereo reconstruction. The two cameras are calibrated beforehand and we also obtained global scale by capturing a checkerboard in the direct and reflected area of the camera image. This provides per-pixel ground truth depth at the main camera. 
In addition, we create waves by perturbing the water surface to evaluate the method's robustness to reasonably large waves (\figref{fig:quant_eval}, the second row).

\figref{fig:quant_eval}(c) and (d) show that our method recovers 3D geometry from the single input image (a) with sufficient accuracy as can be seen in the error maps (e) computed against the ground truth. The average depth errors are 6.3\% and 3.2\% for the calm water surface and the wavy water surface, respectively. Note how dark the reflected observations of the scene in the input images are, which renders any simple appearance adaptation for direct stereo reconstruction impossible. The results show that our method successfully extracts the true scene radiance from the reflected observation which enables robust matching against the direction observation for both cases. The recovered scene appearance, including the saturated intensities of the cup in \figref{fig:quant_eval}(b), also show that our method successfully reconstructs high-dynamic range radiance. The missing areas in the reconstruction are occluded from the camera. The simultaneously reconstructed waves shown as an exaggerated surface normal map in \figref{fig:quant_eval}(f), also look reasonable, although there are no means to know the ground truth, suggesting successful disentanglement of scene and wave structures.

\begin{figure}[t]
 \centering
 \includegraphics[trim={0 1cm 0 0},clip ,width=\linewidth]{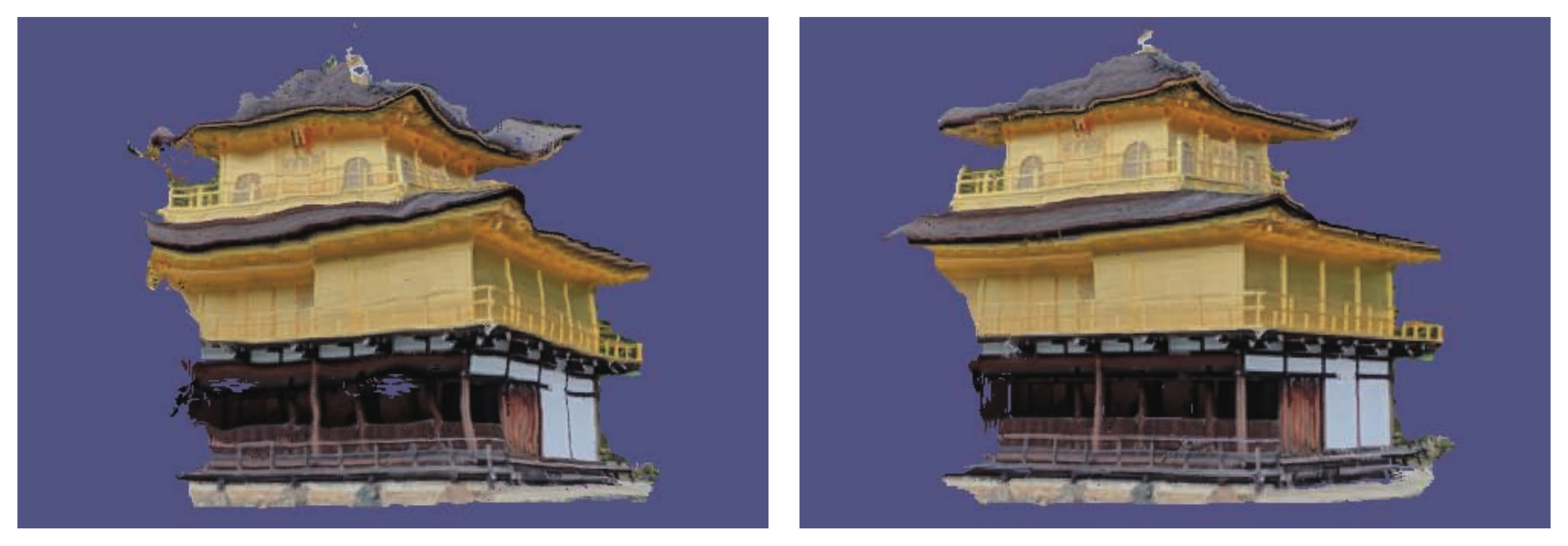}
 \caption{The effect of wave surface recovery: reconstructed 3D model without (left) and with (right) wave geometry estimation. By explicitly recovering the wave surface, our method achieves accurate scene geometry recovery from wavy water reflection.}
 \label{fig:fig_wave_effects}
\end{figure}
To quantitatively evaluate the accuracy of self-calibration, \ie, intrinsic parameter recovery, we randomly sampled $N$ pairs of corresponding point pairs $\VEC{u}$ and $\VEC{u}'$ from the input image in the first row of \figref{fig:quant_eval}(a), and recovered the focal length using the method described in \secref{sec:intrinsics}. We found that with more pairs the relative error decreased rapidly and with $100$ pairs it already reached less than $5\%$ error. Note that, since our method achieves dense matching in the process of radiometry and geometry reconstruction, many more than $100$ point pairs can easily be furnished. 
In \figref{fig:estimate_intrinsics}, we also quantitatively demonstrate the effect of self-calibration. The results clearly show that the focal length estimation undoes the skew and results in more visibly accurate geometry.

\begin{figure*}[t]
 \centering
 \includegraphics[width=\linewidth]{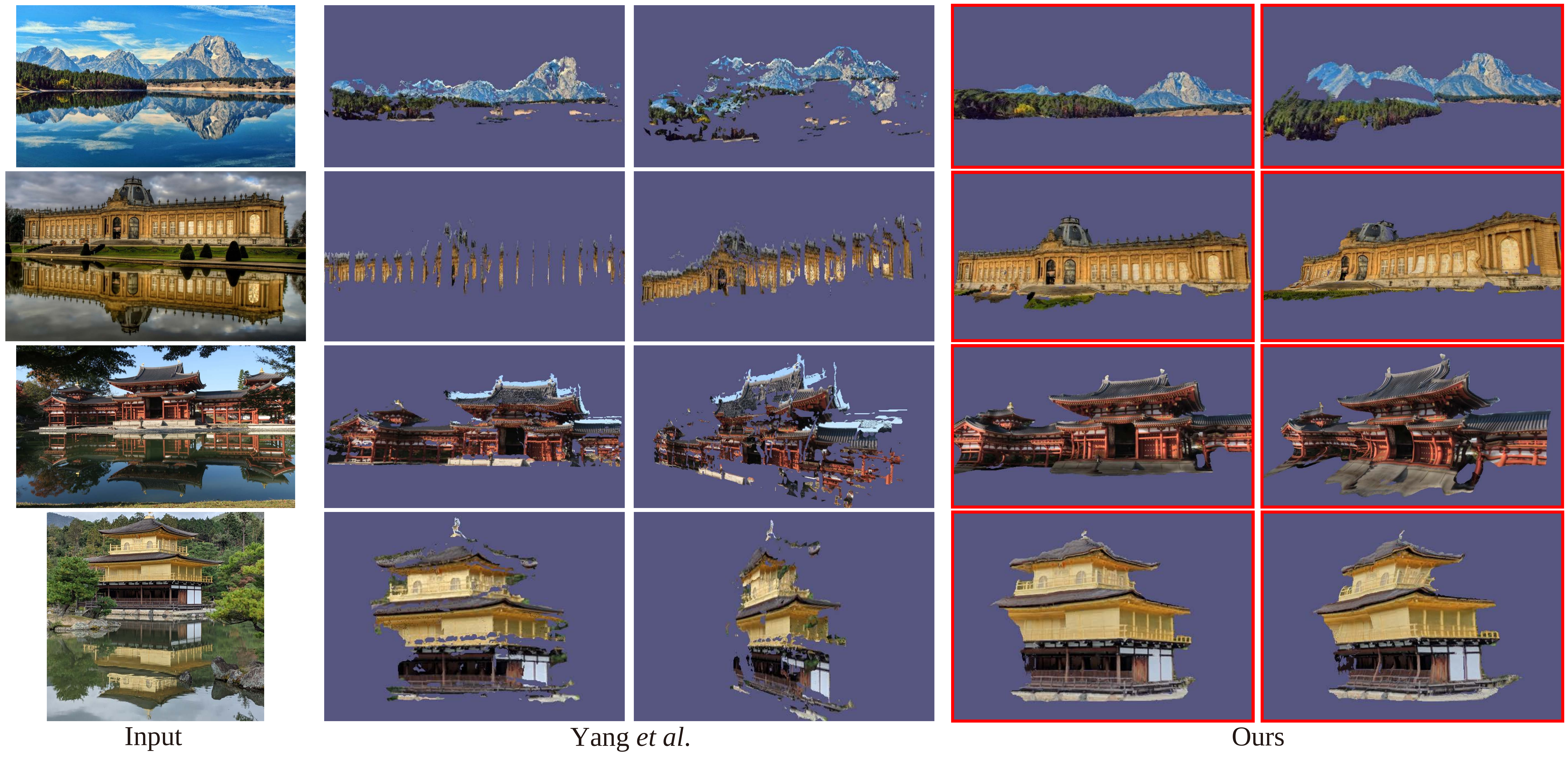}
 \caption{Comparison with Yang \etal\cite{YangTIP_15}. From left to right, input images, results by Yang \etal\cite{YangTIP_15}, and results by our method. Our method successfully reconstructs a dense accurate 3D model of the scene, together with its high-dynamic range appearance. (Image credits for the third and fourth input images: Yuji Wada and Ko Nishino.)}
 \label{fig:comparison}
\end{figure*}

\begin{figure*}[t]
 \centering
 \includegraphics[trim={0 0 0 0.5cm},clip ,width=\linewidth]{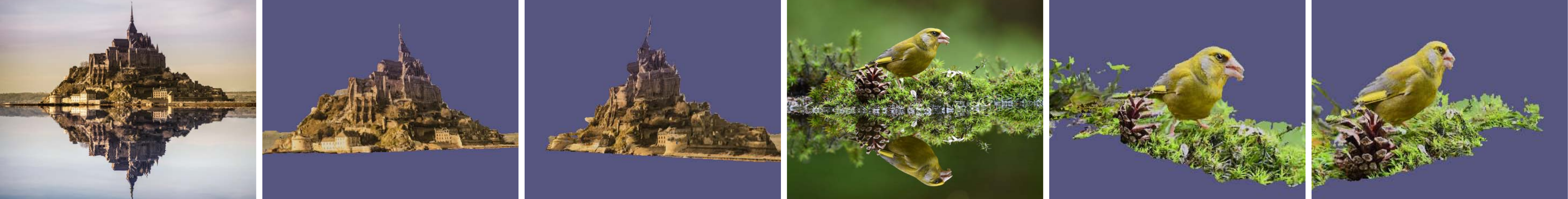}
 \caption{Results on arbitrary images found on the Internet. For each example, we recover an HDR-texture mapped mesh model shown from two different viewing directions. More results can be found in the supplementary video. The results demonstrate the effectiveness of our method on arbitrary already-taken images. (Image credits for the left and right input images: Andr\'{e}s Nieto Porras and Edwin Giesbers.)}
 \label{fig:results}
\end{figure*}
\figref{fig:fig_wave_effects} shows results of reconstructing the Golden Temple with and without estimation of the waves on the pond reflecting it. The results clearly show that by explicitly modeling the waves as 2D deformations of the proxy plane and imposing natural priors on their shape, they can be disentangled from the target geometry.

\figref{fig:comparison} shows comparisons of our method and our implementation of \cite{YangTIP_15}. We compare on two images taken from \cite{YangTIP_15} and two images we have captured. The results clearly show that our method achieves more accurate geometry estimation, in addition to the fundamental difference of also recovering HDR appearance. In general, the recovered geometry by the method in \cite{YangTIP_15} is fragmented, which is a direct result of being inherently restricted to very sparse depth layers due to the heavy reliance on sparse keypoint correspondence pairs that governs the depth range and appearance adaptation. In sharp contrast, by disentangling the radiometric and geometric properties of water reflection, our method is able to achieve dense and clean per-pixel geometry reconstruction.

We apply our method to various images either taken by our phone cameras or found on the Internet. \figref{fig:results} shows the single input images and recovered 3D models with high-dynamic range appearance. The supplementary video contains more results. The results show that the 3D scene structure can be recovered despite waves and complex water surface reflection. It is also interesting to see how the method applies to a wide range of scene scales, ranging from a small bird to a large architecture. Images taken with long focal length tend to result in flatter surface with limited depth variation as one expects. The results also include various types of water reflection, ranging from a puddle to a lake demonstrating its successful application to images truly taken in the wild.
%





\section{Conclusion}
In this paper, we introduced appearance and shape from water reflection to recover 3D geometry and high-dynamic range appearance of a scene from an image capturing both direct and water-reflected views in a single exposure. The method can recover camera parameters and waves in addition to the scene structure and appearance, enabling its application to unconstrained, already-taken images. Experimental results demonstrate its robustness to waves and its effectiveness when applied to arbitrary images taken in the wild. We believe the method has strong implications in a wide range of domains, not just in vision and graphics, but also in photography as a new visual media, as well as in image forensics analysis in which direct and water-reflected geometry and radiometry, even with waves, can now be used as visual cues of image tampering. 

\paragraph{Acknowledgement} This work was in part supported by JSPS KAKENHI 15H05918, 17K20143, 18K19815, and 26240023.

\clearpage
{\small
\bibliographystyle{ieee}
\bibliography{sfwr_wacv_cameraready}
}

\end{document}